\documentclass[times,twocolumn,final,authoryear]{elsarticle}

\usepackage{prletters}
\usepackage{framed,multirow}

\usepackage{amssymb}
\usepackage{latexsym}

\usepackage[utf8]{inputenc}
\usepackage{enumitem}
\usepackage{url}
\usepackage{booktabs}
\usepackage{tabularx}
\usepackage{makecell}
\newcolumntype{C}{>{\centering\arraybackslash}X}
\usepackage{xcolor}
\usepackage{tikz}
\usetikzlibrary{arrows,positioning,shapes.geometric}
\usepackage{amsmath}
\usepackage{pifont}
\newcommand{\cmark}{\ding{51}}
\newcommand{\xmark}{\ding{55}}
\definecolor{newcolor}{rgb}{.8,.349,.1}

\journal{}

\begin{document}

\ifpreprint
  \setcounter{page}{1}
\else
  \setcounter{page}{1}
\fi

\begin{frontmatter}

\title{Multimodal grid features and cell pointers for Scene Text Visual Question Answering}

\author[1]{Lluís \snm{Gómez}}
\author[1]{Ali Furkan \snm{Biten}}
\author[1]{Rubèn \snm{Tito}}
\author[1]{Andrés \snm{Mafla}}
\author[1]{Marçal \snm{Rusiñol}}
\author[1]{Ernest \snm{Valveny}}
\author[1]{Dimosthenis \snm{Karatzas}}


\address[1]{Computer Vision Center -- Universitat Autònoma de Barcelona, Edifici O, Campus UAB, 08193 Bellaterra (Cerdanyola), Barcelona.}

\received{1 May 2013}
\finalform{10 May 2013}
\accepted{13 May 2013}
\availableonline{15 May 2013}
\communicated{S. Sarkar}

\begin{abstract}

This paper presents a new model for the task of scene text visual question answering, in which questions about a given image can only be answered by reading and understanding scene text that is present in it. The proposed model is based on an attention mechanism that attends to multi-modal features conditioned to the question, allowing it to reason jointly about the textual and visual modalities in the scene. The output weights of this attention module over the grid of multi-modal spatial features are interpreted as the probability that a certain spatial location of the image contains the answer text the to the given question. Our experiments demonstrate competitive performance in two standard datasets. Furthermore, this paper provides a novel analysis of the ST-VQA dataset based on a human performance study.

\end{abstract}



\end{frontmatter}


\section{Introduction}
\label{sec:intro}


For an intelligent agent to answer a question about an image, it needs to understand its content. Depending on the question, the visual understanding skills required will vary: object/attributes recognition, spatial reasoning, counting, comparing, use of commonsense knowledge, or a combination of any of them. Reading is another skill that can be of great use for Visual Question Answering (VQA) and has not been explored until recently by \cite{biten2019scene} and \cite{singh2019towards}. 

Scene text VQA is the task of answering questions about an image that can only can be answered by reading/understanding scene text that is present in it. An interesting property of this task over standard VQA is that the textual modality is present both in the question and in the image representations, and thus calls for a different family of composed models using computer vision (CV) and natural language processing (NLP).

\begin{figure}[t]
    \begin{tabular}{@{}ll@{}} 
    \includegraphics[width=0.47\linewidth]{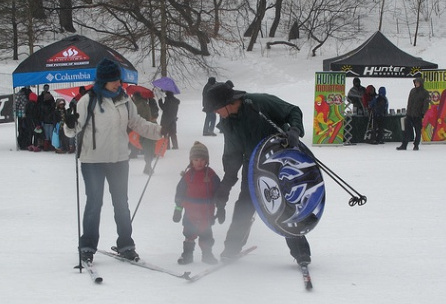} & \includegraphics[width=0.47\linewidth]{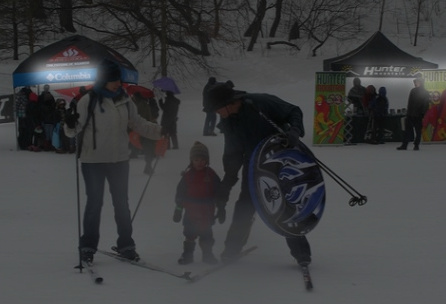} \\
    \multicolumn{2}{@{}l}{\footnotesize{\fontfamily{qhv}\selectfont \textbf{Q:} What brand name is on the tent with the blue stripe?} \par {\color{blue}\footnotesize{\fontfamily{qhv}\selectfont \textbf{A:} COLUMBIA}}}
    \end{tabular}
    \caption{Answering scene text visual questions requires reasoning about the visual and textual information. Our model is based on an attention mechanism that jointly attends to visual and textual features of the image.}
    \label{fig:first}
\end{figure}

\begin{figure*}
    \centering
    \includegraphics[width=\textwidth]{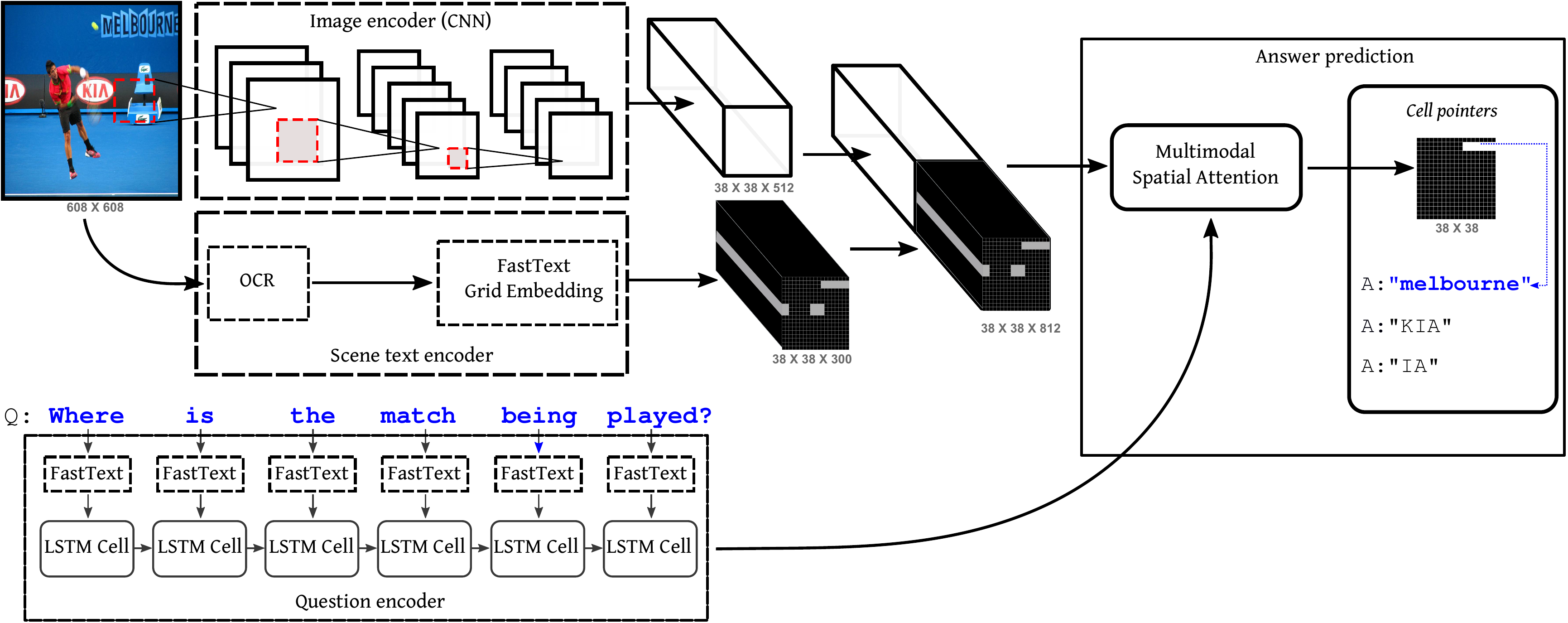}
    \caption{Our scene text VQA model consists in four different modules: a visual feature extractor (CNN), a scene text feature extractor (OCR + FastText), a question encoder (LSTM + FastText), and the answer prediction model.}
    \label{fig:model}
\end{figure*}

Current state of the art on scene text VQA, \cite{singh2019towards}, make use of a dual attention mechanism: one attention module that attends the image visual features conditioned to the question, and another that attends to the textual features (OCR text instances) conditioned to the question. A potential issue with this dual attention is that it makes difficult for the model to reason jointly about the two modalities, since this can only be done after the late fusion of the outputs of the two attention modules. In this paper we propose a solution to this problem, by using a single attention model that attends to multi-modal features as illustrated in Figures~\ref{fig:first} and \ref{fig:model}.

For doing that we construct a grid of multi-modal features by concatenation of convolutional features and a spatial aware arrangement of word embeddings, so that the resulting grid combines the features of the two modalities at each spatial location (cell) of the grid. Then we use an attention module that attends to the multi-modal spatial features conditioned to the question, and the output weights of the attention module are interpreted as the probability that a certain spatial location (grid cell) of the image contains the answer to the given question.

It is worth noting that with such an approach we somehow recast the problem of scene text VQA as an answer localization task: given an image and question our model localizes the bounding box of the answer text instance. In this sense the architecture of our model is similar to single shot object detectors, e.g. \cite{redmon2016you} and \cite{liu2016ssd}, but conditioning their output to a given question in natural language form through an attention layer. Notice this idea also directly links with the pointer networks proposed by \cite{vinyals2015pointer} and used in \cite{singh2019towards}, but distinctly 
to these works,
we have a fixed input-output space: the number of grid cells. 

Another important difference of our model with current state of the art in both standard VQA and scene text VQA is that we use grid based features for encoding the image, while most current models make use of region based features as in \cite{anderson2018bottom}. Although our motivation here is our belief that visual and textual features must be fused together maintaining their spatial co-relation,  this has also other benefits, as the whole model is simplified and the times for training and inference is highly reduced.


\section{Related Work}
\label{sec:related}

Scene text visual question answering has been proposed recently with the appearance of two datasets, TextVQA by \cite{singh2019towards} and ST-VQA by \cite{biten2019scene}. 
The ST-VQA dataset comprises $23,038$ images and $31,791$ question/answer pairs. The images were collected from seven different public data sets with the only requirement to contain at least 2 text tokens, so there is always some inherent confusion. 
The annotation process was carried out by human annotators who received specific instructions to ask questions based on the text present in each image, so that the answer to the questions should always be a token (or a set of tokens) of legible text in the image.

The TextVQA dataset comprises a total of $28,408$ images and $45,336$ questions. 
All images come from the OpenImages dataset, \cite{OpenImages}, and were sampled on a category basis, emphasizing categories that are expected to contain text. In TextVQA any question requiring reading the image text is allowed, including questions for which the answer does not correspond explicitly to a legible text token (e.g. binary (yes/no) questions). Notice that distinct from ST-VQA answering those questions implies the use of a fixed output vocabulary.

In parallel to these works~\cite{mishra2019ocr} presented the OCR-VQA dataset, with more than 1 million question-answer pairs about 207K images of book covers. 
However, the task in this dataset is different in nature to the one our model is designed for, since more than $50\%$ of the questions have answers that are not scene text instances (including $40\%$ binary (yes/no) questions and $10\%$ questions about book genres).

Along with the ST-VQA dataset, \cite{biten2019scene} presented a baseline analysis including standard VQA models by \cite{kazemi2017show} and \cite{yang2016stacked}, and a variation of those models in which image features where concatenated with a text representation obtained with a scene text retrieval model \cite{gomez2018single} that produces a PHOC representation on its output. Our model takes inspiration from this concatenation of visual and textual features along the spatial dimensions, but we replace the PHOC structural descriptor by semantic word embeddings.

\cite{biten2019icdar} organized the ICDAR 2019 Competition on Scene Text Visual Question Answering, in which a total of seven teams evaluated their models on the ST-VQA dataset. The winner entry (VTA) was based on the Bottom-Up and Top-Down VQA model by \cite{anderson2018bottom} but the textual branch was enhanced with BERT word embeddings, \cite{devlin2018bert}, of both questions and text instances extracted with an off-the-shelf OCR system.

\cite{mishra2019ocr} presented a model that represents questions using a BLSTM, images using a pretrained CNN, and OCRed text with their average word2vec representations. They encode each OCRed text block (a group of text tokens) using its coordinate positions, and a semantic tag provided by a named entity recognition model. All these features 
are  concatenated and fed into a MLP network that predicts an answer from a fixed vocabulary (including ``yes'', ``no'', and 32 predefined book genres) or from one of the OCRed text blocks.

On the Text-VQA side, \cite{singh2019towards} proposed the Look, Read, Reason \& Answer (LoRRA) method, that extends the well known framework for VQA of \cite{singh2018pythia} by allowing to copy an OCR token (text instance) from the image as the answer. For this they apply an attention mechanism, conditioned on the question, over all the text instances provided by the OCR model of \cite{borisyuk2018rosetta}, and include the OCR token indices as a dynamic vocabulary in the answer classifier’s output space. The model uses two attention modules, one attends the visual features 
and the other attends to textual features, both conditioned on the question. After that the weighted average over the visual and textual features are concatenated and go through a two-layer feed-forward network which predicts the binary probabilities as logits for each answer.

\cite{singh2019towards} have also organized the TextVQA Challenge 2019, in which the winner method (DCD\_ZJU) extended the LoRRA model by using the BERT embedding instead of GloVE, \cite{pennington2014glove}, and the Multi-modal Factorized High-order (MFH) pooling proposed by \cite{yu2018beyond} in both of the attention branches. 


The main difference of the model proposed here with the LoRRA and DCD\_ZJU models is that we use a single attention branch, that attends jointly to visual and textual features. 
We also use a different pointer mechanism that directly treats the output weights of the attention module as the probability that a certain cell contains the correct answer to a given question. Notice that this is closer to the original formulation of Pointer Networks \cite{vinyals2015pointer} since we directly use the predicted weights of the attention module as pointers, without any extra dense layer as in \cite{singh2019towards}, but slightly different in the sense that our input and output size is fixed by the size of the features' grid. On the other hand in our model we use a single shot object detector as a visual feature extractor instead of the Faster-RCNN used in LoRRA, which implies faster training and inference times.







\section{Method}
\label{sec:method}
Figure \ref{fig:model} illustrates the proposed model, it consists in four different modules: image encoder (CNN), scene text encoder (OCR + FastText), question encoder (LSTM + FastText), and the answer prediction module. The CNN, OCR, and FastText models are used with pre-trained weights and not updated during training, while the question encoder and answer prediction modules are trained from scratch.

\subsection{Image encoder}

One common component of all visual question answering models is the use of a convolutional neural network as a visual feature extractor. While in the first VQA models it was common to use a single flat vector as a global descriptor for the input image, see \cite{antol2015vqa} and \cite{kim2016multimodal}, with the advent of attention mechanisms grid based features became ubiquitous, see e.g. in \cite{kazemi2017show} and \cite{yang2016stacked}. However, today's standard approach is to use region based convolutional features from a set of objects provided by an object detection network as proposed in \cite{anderson2018bottom}. The rationale is that using objects as the semantic entities for reasoning helps for a better grounding of language.

In this paper we are interested in using grid features, because our whole motivation depends on them. But contrary to previous models using grid features, we propose here to extract them using a single shot object detector, \cite{redmon2018yolov3}, instead of CNN models pretrained for classification. With this we argue that it is possible to maintain a fair trade-off between the use of objects' representations for reasoning and the spatial structure of the grid-based features.


Our visual feature extraction $f_{CNN}(I)$ is based on the architecture of the YOLOv3 model by \cite{redmon2018yolov3} with weights pre-trained on the MS-COCO dataset. The YOLOv3 model has a total of 65 successive $3 \times 3$ and $1 \times 1$ convolutional layers and residual connections. 
We extract features from the $61$st layer, which produces  a feature map with dimensions $38 \times 38 \times 512$ that encode high-level object semantics 
This configures the features' grid size in our model to be $38 \times 38$. The size of the grid is chosen so that we can quantize the textual information without loosing small words (see next section). A $38 \times 38$ grid size means each cell corresponds to a $16 \times 16$ patch of the input image (with an $608 \times 608$ resolution), which means the smallest possible bounding box of a text instance we expect to find is $16 \times 16$.


\subsection{Scene text encoder}
\label{sec:text}

The first step in our textual feature extractor $f_{ST}(I)$ is to apply an optical character recognition (OCR) model to the input image in order to obtain a set of word bounding boxes and their transcriptions $T~=~\{(b_1, t_1),~(b_2, t_2), \dots,~(b_n, t_n)\}~$. Text extraction from scene images is still an open research area attracting a lot of interest among the computer vision research community, see e.g. \cite{Baek_2019_ICCV,liu2018fots,buvsta2018e2e}. In this work we have evaluated several publicly available state of the art models as well as the commercial OCR solution of Google\footnote{\url{https://cloud.google.com/vision/}}.

As a standard practice in many applications of natural language processing we embed the words extracted from the OCR module into a semantic space by using a pretrained word embedding model. In our case we make use of the FastText word embedding by \cite{bojanowski2016enriching}, because it allows us to embed out of vocabulary (OOV) words. Notice that OOV words are quite common in scene text VQA because of two reasons: first, some question may refer to named entities or structured textual information that is not present in closed vocabularies, e.g. telephone numbers, e-mail addresses, website URLs, etc.; second, the transcription outputs of the OCR may be partially wrong, either because the scene text is almost illegible, partially occluded or out of the frame. 

We use the FastText pretrained model 
with 1 million $300d$ word vectors, trained with subword information on Wikipedia 2017, UMBC webbase corpus and statmt.org news dataset.

\begin{figure}[h]
    \centering
    \begingroup
    \setlength{\tabcolsep}{1pt} 
    \renewcommand{\arraystretch}{0.75} 
    \begin{tabular}{@{}c c c@{}}
        \includegraphics[width=0.325\linewidth]{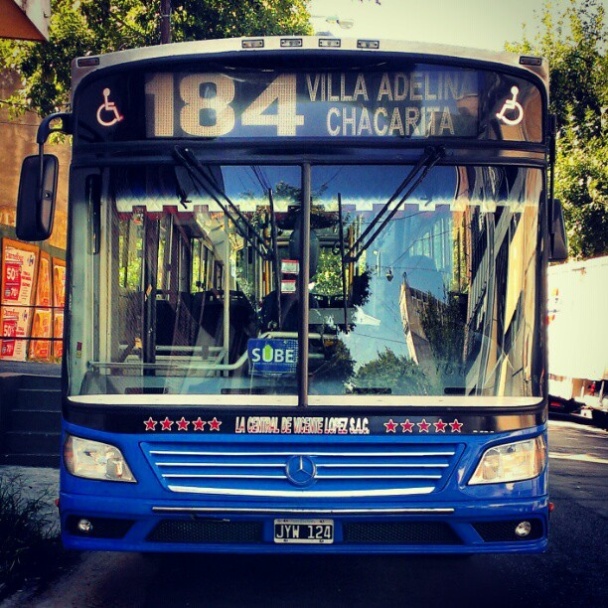} & \includegraphics[width=0.325\linewidth]{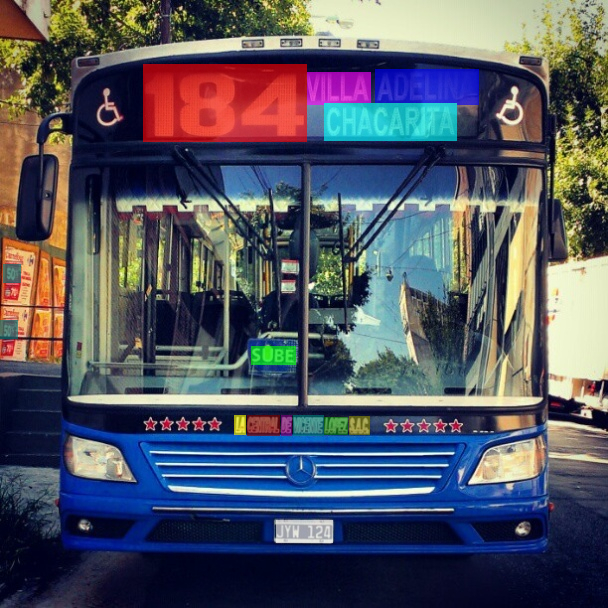} & \includegraphics[width=0.325\linewidth]{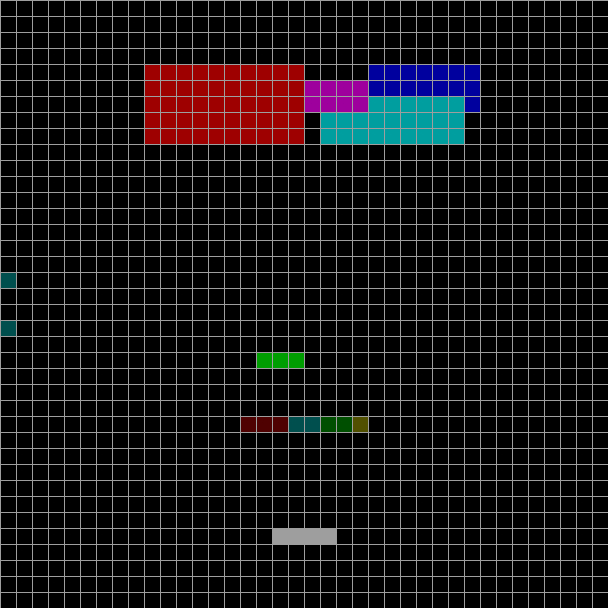}   \\
        (a) & (b) & (c)  
    \end{tabular}
    \endgroup
    
    \caption{Grid cell assignment of the OCR words' bounding boxes. Given an input image (a), the bounding boxes of the words extracted from the OCR model (b) are assigned to their overlapping cells.}
    \label{fig:ocr_grid}
\end{figure}

With all word transcriptions in $T$ embedded in the FastText $300d$ space we construct a $38 \times 38 \times 300$ tensor by assigning each of their bounding boxes to the cells in a $38 \times 38$ grid with which they overlap as illustrated in Figure \ref{fig:ocr_grid}, so that the embedding vectors maintain the same relative spatial positions as the words in the original image. In order to overcome small words being overlapped by larger words we do this assignment in order, from larger words to smaller. The cells without any textual information are set to zero value. Finally, we concatenate the outputs of the image encoder and the scene text encoder to obtain the multi-modal grid based features of the image $f_m(I) = [f_{CNN}(I); f_{ST}(I)] \in R^{38 \times 38 \times 812}.$  

\subsection{Question encoder}

The question encoder is another common module in all VQA models. Recurrent neural networks, either with LSTM or GRU units, are the most common choice of state of the art models, e.g. \cite{yang2016stacked} \cite{kazemi2017show} \cite{jiang2018pythia} \cite{anderson2018bottom} \cite{singh2019towards}, while the use of CNN has also been explored as an alternative encoding in \cite{yang2016stacked}. In this work we use an LSTM encoder, with the LSTM unit formulation 
of \cite{gers1999learning}.



Given a question $Q$ with $N$ words $Q = \{q_1, q_2, \dots, q_N\}$ we first embed each word with the FastText word embedding function described in section~\ref{sec:text}, and then we feed each word embedding vector into the LSTM. The final hidden layer of the LSTM model is taken as the output of the question encoder:

\begin{equation}
    f_q(Q) = LSTM({\tilde q}_i, h_{i-1}) \forall i \in \{1,2,\dots,N\}
\end{equation}



\noindent where ${\tilde q}_i$ is the FastText embedding of word $q_i$, and $h_{i-1}$ is the output of the LSTM for previous word -- we omit the propagation of memory units to simplify the notation. Our LSTM has two dense layers with $256$ hidden units and two Dropout layers with a $0.5$ drop out rate. The output of the question embedding function $f_q(Q)$ is a vector with $1024$ dimensions.

\subsection{Answer prediction}
\label{sec:answer_pred}
The main component of the answer prediction module is an attention mechanism that attends to the spatial multi-modal features $f_m(I)$ conditioned on the question embedding $f_q(Q)$. 



\begin{figure}[h]
    \centering

    \begin{tikzpicture}[>=latex',scale=0.75, every node/.style={scale=0.75}]
        \tikzset{block/.style= {draw, rectangle, align=center,minimum width=1cm,minimum height=0.5cm},
        }
        \tikzset{var/.style= {align=center,minimum width=2cm,minimum height=0.5cm},
        }
        \tikzset{op/.style= {draw, circle, align=center,minimum width=0.5cm,minimum height=0.5cm},
        }
        
        \node [var]  (fv) {$f_m(I)$};
        \node [var, right = 1.6cm of fv] (fq){$f_q(Q)$};
        \node [block, above = 0.5cm of fv] (conv1){$1 \times 1$ Conv};
        \node [block, above = 0.25cm of conv1] (tanh1){tanh};
        \node [block, above = 0.5cm of tanh1] (conv2){$1 \times 1$ Conv};
        \node [block, right = 2.cm of conv1] (dense){Dense};
        \node [block, above = 0.25cm of dense] (tanh3){tanh};
        \node [block, above = 0.5cm of tanh3] (tile1){Tile};
        \node [op, above right = 0.31cm and 0.79cm of conv2](add){$+$};
            \node [coordinate, above = 0.47cm of conv2] (cord1){};
        \node [coordinate, above = 0.49cm of tile1] (cord2){};
        \node [block, above = 0.25cm of add] (tanh4){tanh};
        \node [block, above = 0.25cm of tanh4] (conv3){$1 \times 1$ Conv};
        \node [block, above = 0.25cm of conv3] (sigmoid){$\sigma$};
        \node [var, above = 0.5cm of sigmoid] (p){$p_{att}$};
        
        \path[draw, ->]
            (fv) edge (conv1)
            (conv1) edge (tanh1)
            (tanh1) edge (conv2)
            (conv2) -- (cord1)
            (cord1) edge (add)
            (add) edge (tanh4)
            (tanh4) edge (conv3)
            (conv3) edge (sigmoid)
            (sigmoid) edge (p)

            (fq) edge (dense)
            (dense) edge (tanh3)
            (tanh3) edge (tile1)
            (tile1) -- (cord2) 
            (cord2) edge (add)
                    ;

    \end{tikzpicture}
    \caption{Computation graph of our attention mechanism $f_{Att}$.}
    \label{fig:att_graph}
\end{figure}
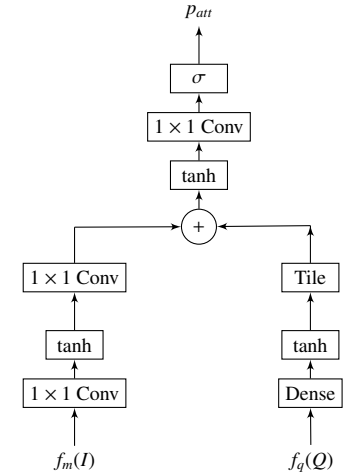

Figure \ref{fig:att_graph} illustrates the computation graph of our attention mechanism $f_{Att}$. First the multimodal grid features $f_m(I)$ are convolved by two $1 \times 1$ convolutional layers with 1024 and 512 kernels respectively, resulting in a $38 \times 38 \times 512$ tensor, the question encoded vector $f_q(Q)$ goes through a dense layer with $512$ output neurons and is tiled/broadcasted to a shape of $38 \times 38 \times 512$. These two tensors ($m_{att}$ and $q_{att}$) are added and activated with an hyperbolic tangent (tanh) activation. Finally, the resulting tensor of this operation is convolved with a $1 \times 1$ convolutional layer with a sigmoid activation function to produce the output attention map $p_{att}$ with shape $38 \times 38 \times 1$:

\begin{equation}
    p_{att} = f_{Att}([f_{CNN}(I); f_{ST}(I)], f_q(Q))
\end{equation}

At this point we interpret the values in the output attention map $p_{att}$ as the probability of each image cell to contain the correct answer to the given question $Q$. Notice that by applying a sigmoid activation function to the last convolution layer 
we treat the probability for each cell as an individual binary classification problem. This is intentional as in most of the cases the bounding box of the correct answer will cover more than one cell. We train our model using the binary cross entropy loss function:

\vspace{-2em}
\begin{equation}
    E = - \sum\limits_{i=1}^{38} \sum\limits_{j=1}^{38} \left[ g_{i,j} \log p_{i,j} + (1 - g_{i,j}) \log(1 - p_{i,j}) \right]
\end{equation} 

\noindent where $p_{i,j}$ is the probability value of the cell on the $i$th row and $j$th column on the output attention map $p_{att}$, and $g_{i,j}$ is the ground truth value for that cell: $1$ if the cell contains the answer, $0$ otherwise. 
At inference time, the predicted answer is the OCR token assigned to the cell with maximum probability.

The attention mechanism described so far can be used within several design variations such as the stacked attention of \cite{yang2016stacked}, or the question-image co-attention of \cite{lu2016hierarchical} and \cite{nam2017dual}. In particular we have adopted the stacked design in our model and empirically found an improvement over using a single attention layer (see the ablation study in section \ref{sec:ablation} for the details). For this we stack two attention layers, and in the first one we combine the weighted average over the multimodal spatial features (using the output probability map as weights) with the question embedding (by addition), and this combination is fed to the second attention layer as the question embedding.

Moreover, we notice that since our model is made fully convolutional (including the image encoder) on all the visual branch, we can perform inference at different input scales using the same learnt weights. 




\begin{figure*}[t]
\begin{center}
\begin{tabular}{p{0.17\textwidth} p{0.17\textwidth} p{0.17\textwidth} p{0.17\textwidth} p{0.17\textwidth}}

\includegraphics[width=\linewidth,height=\linewidth]{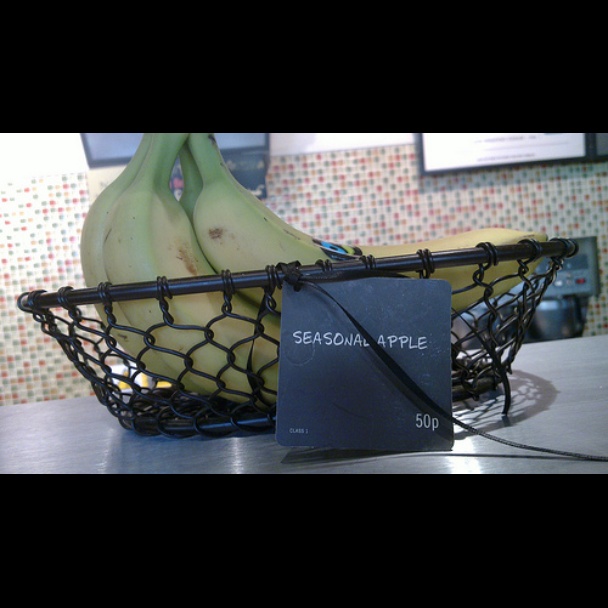}& 
\includegraphics[width=\linewidth,height=\linewidth]{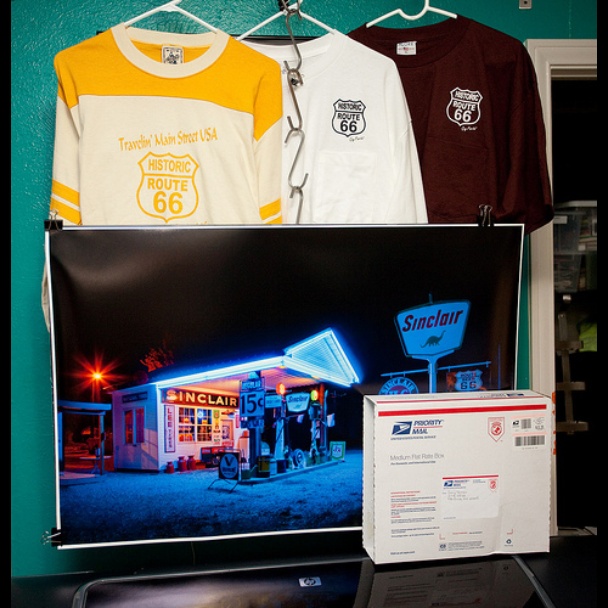}& 
\includegraphics[width=\linewidth,height=\linewidth]{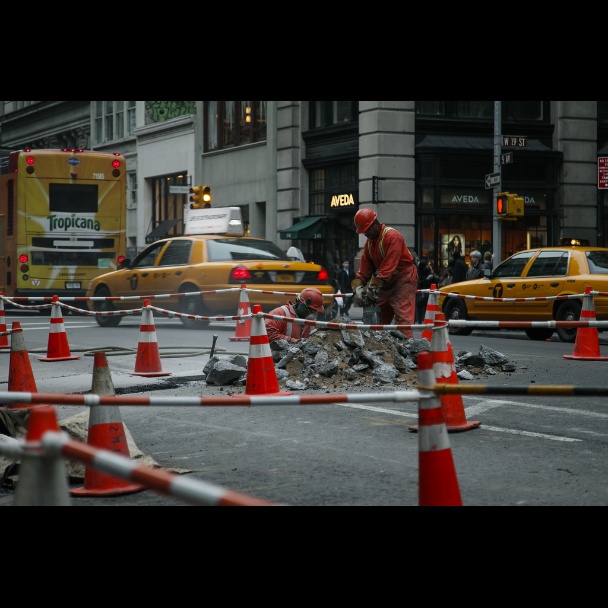}& 
\includegraphics[width=\linewidth,height=\linewidth]{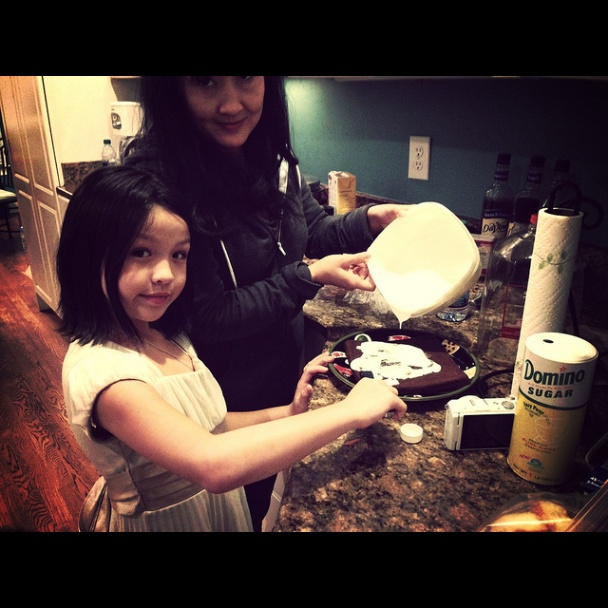}& 
\includegraphics[width=\linewidth,height=\linewidth]{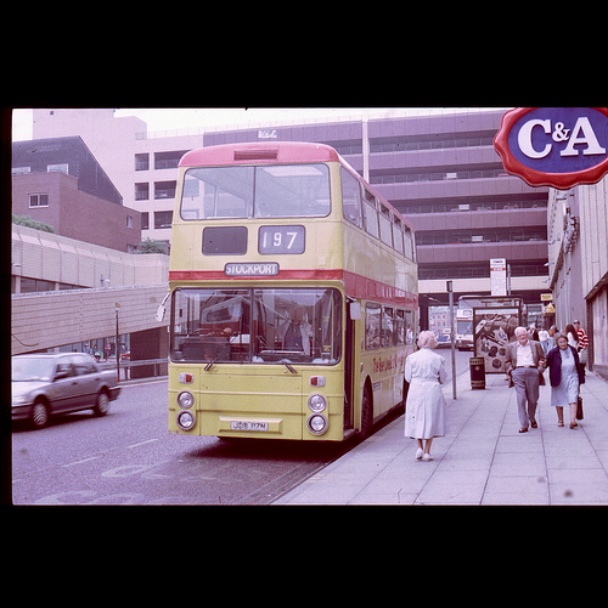}\\ 

\includegraphics[width=\linewidth,height=\linewidth]{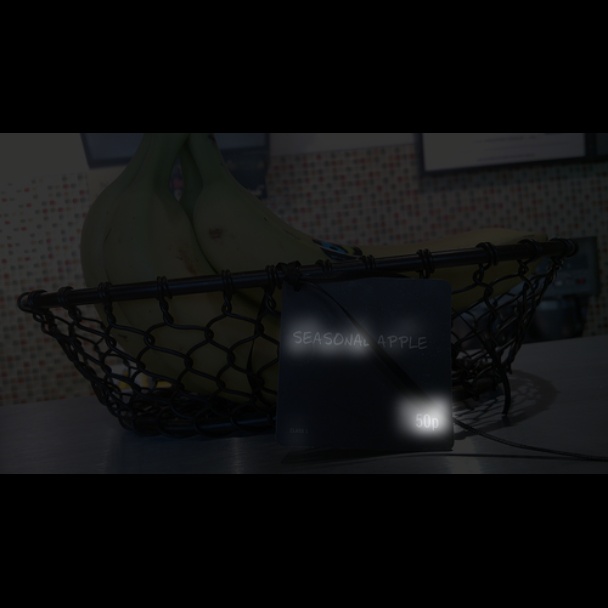}& 
\includegraphics[width=\linewidth,height=\linewidth]{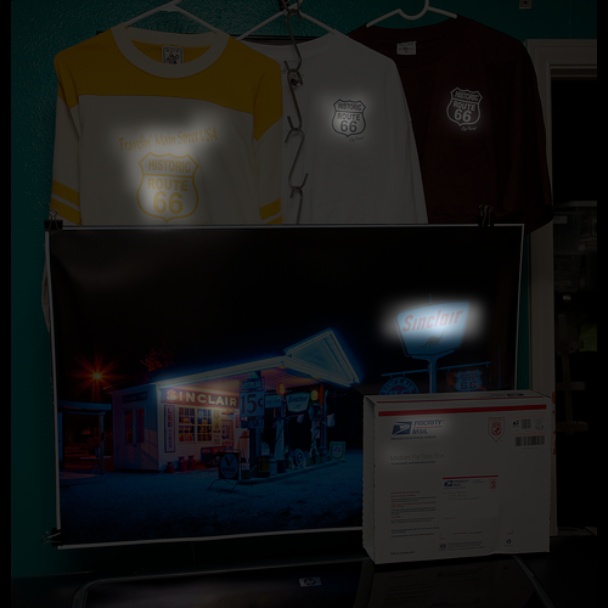}& 
\includegraphics[width=\linewidth,height=\linewidth]{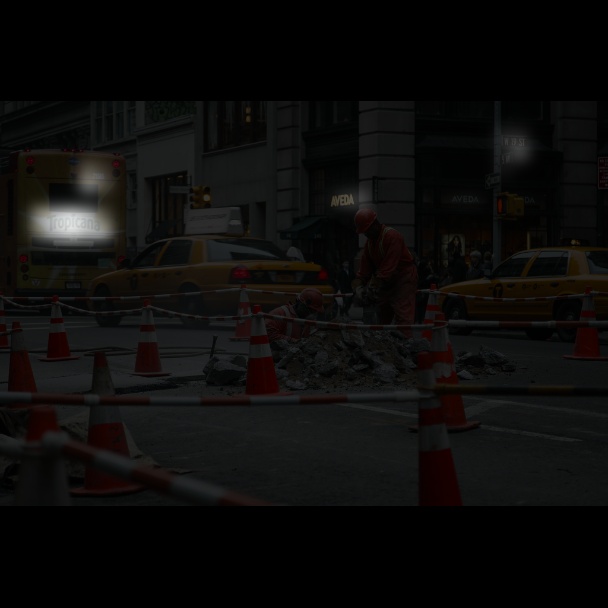}& 
\includegraphics[width=\linewidth,height=\linewidth]{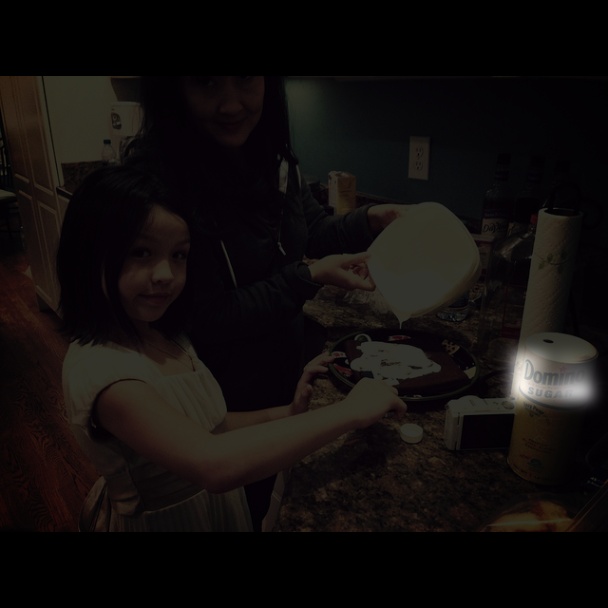}& 
\includegraphics[width=\linewidth,height=\linewidth]{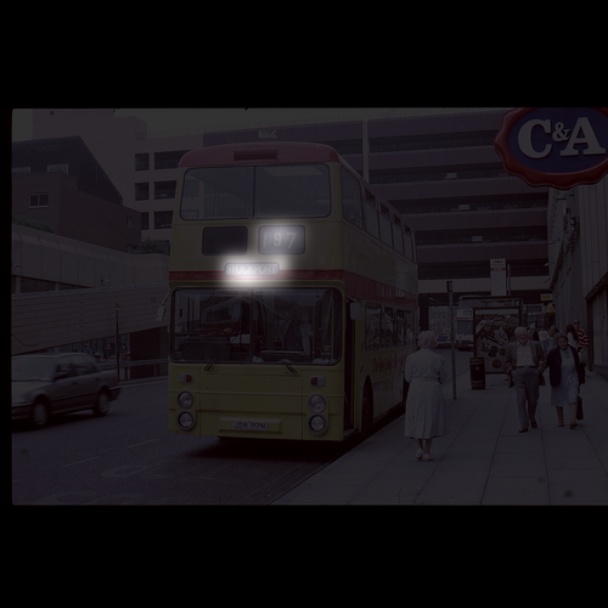}\\ 

\footnotesize{\fontfamily{qhv}\selectfont \textbf{Q:} How much is the rate of one seasonal apple fruit?} \par{\color{blue}\footnotesize{\fontfamily{qhv}\selectfont \textbf{A:} 50p}}& 
\footnotesize{\fontfamily{qhv}\selectfont \textbf{Q:} What is the word on the white sign?} \par{\color{blue}\footnotesize{\fontfamily{qhv}\selectfont \textbf{A:} sinclair}}& 
\footnotesize{\fontfamily{qhv}\selectfont \textbf{Q:} What is the brand advertised on the back of the bus?} \par{\color{blue}\footnotesize{\fontfamily{qhv}\selectfont \textbf{A:} tropicana}}& 
\footnotesize{\fontfamily{qhv}\selectfont \textbf{Q:} What is the sugar brand?} \par{\color{blue}\footnotesize{\fontfamily{qhv}\selectfont \textbf{A:} domino}}& 
\footnotesize{\fontfamily{qhv}\selectfont \textbf{Q:} What is the destination of the bus ?} \par{\color{blue}\footnotesize{\fontfamily{qhv}\selectfont \textbf{A:} stockport}}
\end{tabular}
\end{center}
\vspace{-6pt}
\caption{Examples of questions from the ST-VQA tests and correctly predicted answers by our model.}
\label{fig:qualitative}
\end{figure*}

\section{Experiments}
\label{sec:experiments}

In this section we present a set of experiments performed on the ST-VQA and TextVQA datasets. First, we briefly introduce the metrics of both datasets and present a comparison of different OCR systems on ST-VQA. Second we compare the performance of the proposed model with the state of the art on both datasets, and present an ablation study of the proposed model. Finally, we present an extension of the ST-VQA dataset and analyze human performance on a subset of its test set.


The evaluation metric on ST-VQA is the average normalized Levenshtein similarity (ANLS) that assigns a soft score $s$ to a given pair of predicted and ground-truth answers ($ans_{pred}$ and $ans_{gt}$) based on their normalized Levenshtein edit distance ($d_{LN}$): $s(ans_{pred}, ans_{gt}) = 1 - d_{NL}(ans_{pred}, ans_{gt})$. On the other hand, the evaluation metric on TextVQA is the VQAv2 accuracy: $Acc(ans)=min(\frac{h(ans)}{3},1)$ where $h(3)$ counts the number of humans that answered $ans$ among the $10$ collected human answers for each question.

Table~\ref{tab:ocr} shows the answer recall of two different state of the art scene text recognition models and of a commercial OCR system. Answer recall is computed as the percentage of answers in the ST-VQA train set that match with a text token found by the OCR system. The ANLS upper-bound gives us the maximum score we can achieve in this dataset with different OCR systems.

\begin{table}[h]
    \centering
    \begin{tabularx}{\linewidth}{X c c}
    \toprule
    OCR  &  \makecell{Answer \\ Recall} & \makecell{ANLS  \\ Upper-bound}\\
    \midrule
    FOTS -- \cite{liu2018fots} & 37.56 & 0.47 \\
    E2EML -- \cite{buvsta2018e2e} & 41.37 & 0.52 \\
    Google OCR API & 60.19 & 0.74 \\
    \bottomrule
    \end{tabularx}
    \caption{Answer recall and ANLS upper-bound for different off-the-shelf OCR systems on the ST-VQA training set.}
    \label{tab:ocr}
\end{table}

For all experiments reported in this section on the ST-VQA dataset we use the OCR tokens obtained with the Google OCR API. For the experiments on the TextVQA dataset we use the OCR tokens from the Rosetta OCR system, \cite{borisyuk2018rosetta}, that are provided with the dataset to showcase comparable results. At training time we discard image/question pairs for which the answer is not in the OCR tokens' set.

\subsection{Performance comparison}


Table~\ref{tab:stvqa} compares the performance of the proposed model with the state of the art on the ST-VQA dataset. We appreciate that our model clearly outperforms all previously published methods both in ANLS and accuracy, improving more than $10\%$ ANLS compared to the ST-VQA competition models and $5\%$ ANLS over LoRRA. It is important also to recall here that our model is $5 \times$ faster than LoRRA at processing an image, as a consequence of using YOLOv3 instead of Faster-RCNN for feature extraction.

\begin{table}[h]
    \centering
    \begin{tabularx}{\linewidth}{X l r}
    \toprule
    Method  &  ANLS & Acc.\\
    \midrule
    SAAA~\cite{kazemi2017show} & 0.087 & 6.66 \\
    SAN~\cite{yang2016stacked} & 0.102 & 7.78 \\ 
    \midrule
    SAN+STR~\cite{gomez2018single} & 0.136 & 10.34 \\ 
    QAQ - rep. from ~\cite{biten2019icdar} & 0.256 & 19.19 \\
    VTA - rep. from ~\cite{biten2019icdar} & 0.282 & 18.12 \\
    LoRRA~\cite{singh2019towards} & 0.331\dag & 21.28 \\
    \bf{Ours} & \bf{0.381} & \bf{26.06} \\
    \bottomrule
    \end{tabularx}
    \caption{ST-VQA performance comparison on the test set. Numbers with~\dag~are from the official implementation of LoRRA trained on ST-VQA using the same OCR tokens as in our model.}
    \label{tab:stvqa}
\end{table}

Figure~\ref{fig:qualitative} shows qualitative examples of the produced attention masks and predicted answers for $5$ image/question pairs from the ST-VQA test set that are correctly answered by our model. Among them we can see examples in which textual information alone would suffice to provide a correct answer, but also cases where a joint interpretation of visual and textual cues is needed. More qualitative examples are provided as supplementary material of this paper.

Table~\ref{tab:textvqa} shows the performance comparison on the validation set of TextVQA. In this case we also compare the accuracy in the specific subset of questions for which the answer is among OCR tokens (indicated as Acc.\dag~in the table), to understand how the presence of answers that do not correspond to scene text instances in the image (e.g. \textit{``yes''/``no''} answers) affect the performance of our model. In this subset our model outperforms previous state of the art by a clear margin, while in the whole validation set we observe the opposite. Notice that this is expected because our model has no mechanism for providing valid answers to questions the answers of which are not in the OCR tokens,
while the LoRRA model can cope with these questions by using a fixed vocabulary answer output space similar to standard VQA models.

\begin{table}[h]
    \centering
    \begin{tabularx}{\linewidth}{X r r}
    \toprule
    Method  &  Acc.\dag & Acc.\\
    \midrule
    SAAA~\cite{kazemi2017show} & 9.09 & 13.33 \\
    \midrule
    LoRRA~\cite{singh2019towards} & 32.03 & \bf{27.48} \\
    \bf{Ours} & \bf{37.60} & 21.88 \\
    \bf{Ours + SAAA} & \bf{37.71} & 26.59 \\
    \bottomrule
    \end{tabularx}
    \caption{TextVQA performance comparison on the validation set. Acc.\dag~refers to the subset of questions with answers among OCR tokens.}
    \label{tab:textvqa}
\end{table}

In order to provide a fair comparison in the whole validation set of TextVQA we have combined the predictions of our model with the well known standard VQA model SAAA, \cite{kazemi2017show}. In this experiment we have trained the SAAA model on TextVQA with a fixed output space of the most common $3,000$ answers, and the results of entry Ours+SAAA in Table~\ref{tab:textvqa} correspond to an ensemble model in which the provided answer is selected from SAAA if the classification confidence is larger than a given threshold or from our model otherwise. In particular, the best threshold experimentally found over validation data is $0.37$. 
We appreciate that this ensemble model achieves competitive performance to the state of the art. While SAAA alone has a marginal performance in TextVQA, the confidences of its predictions are good indicators for whether a given question can be answered without reading the scene text. In such a scenario a model like ours can be leveraged in a mixed dataset where questions may or may not require answers from the OCR tokens' set.

\subsection{Ablation study and effect of different pre-trained models}
\label{sec:ablation}
In this section we perform ablation studies and analyze  the effect of different pre-trained models and off-the-shelf OCR systems to our method's performance.

Table~\ref{tab:ablation} shows ablation experiments for different attention mechanisms in our model. \textbf{FCN} stands for a Fully Convolutional Network in which three convolutional layers (with respectively  $512$, $256$, and $1$ $3 \times 3$ kernels, ReLU activations and Batch Norm) are applied to the concatenation of features from the YOLOv3 model, the grid of OCR tokens' FastText embedding vectors, and the (tiled) LSTM question embedding. This model has no attention mechanism, but produces at its output a $38 \times 38$ grid as in our model and can be trained in the same way. The \textbf{FCN + Dual Att.} model uses a dual attention mechanism similar to the LoRRA model: one attention module attends the YOLOv3 features conditioned to the question, and the other attends to the grid of OCR tokens FastText vectors conditioned to the question. The outputs of those two attention modules are then concatenated and fed into a convolutional block (similar as for the FCN model) to produce the $38 \times 38$ output. Finally, \textbf{FCN + Multi-modal Att.} and \textbf{FCN + Stack Multi-modal Att} correspond to the proposed model, with one and two multi-modal attention layers respectively as explained in section~\ref{sec:answer_pred}. We can point out that the dual attention mechanism is not helping at all under this set-up, while our multi-modal attention layers consistently improve the results of the FCN model.

\begin{table}[h]
    \centering
    \begin{tabularx}{\linewidth}{X r}
    \toprule
    Method  &  ANLS\\
    \midrule
    FCN & 0.319 \\
    FCN + Dual Att. & 0.279 \\
    FCN + Multi-modal Att. & 0.355 \\
    FCN + Stack Multi-modal Att. & \bf{0.381}  \\
    
    \bottomrule
    \end{tabularx}
    \caption{Ablation study using different attention mechanisms in our model.}
    \label{tab:ablation}
\end{table}

In Table~\ref{tab:pretrained} we study the effect of different pre-trained word embedding models and CNN backbones in our method performance.

\begin{table}[h]
    \centering
    \begin{tabularx}{\linewidth}{X C C r}
    \toprule
    CNN  &  Q. Emb.  &  OCR Emb.  &  ANLS\\
    \midrule
    Inception v2  & FastText & FastText & 0.319 \\
    ResNet-152  & FastText & FastText & 0.332 \\
    YOLO v3 & FastText & FastText &  \bf{0.381}  \\
    YOLO v3 & BERT & FastText &  0.327  \\
    YOLO v3 & BERT & BERT &  0.310  \\
    \bottomrule
    \end{tabularx}
    \caption{ST-VQA performance using different pre-trained word embedding models and CNN backbones.}
    \label{tab:pretrained}
\end{table}

We observe that the visual features of the YOLOv3 object detection model yield superior performance when compared with pre-trained features of two well known networks for image classification: InceptionV2, \cite{szegedy2016rethinking}, and ResNet-152, \cite{he2016deep}. Also in Table~\ref{tab:pretrained} we appreciate that the FastText pre-trained word embedding works better than the BERT embedding for both the question and OCR tokens' encoders. 





\subsection{ST-VQA extensions and human performance analysis}

With this paper we are releasing an updated version of the ST-VQA dataset that includes the OCR tokens used in all our experiments. This way we make sure any methods using OCR tokens and evaluating in this dataset can be fairly compared under the same conditions. 
Moreover, in order to understand the nature of the dataset better, we have conducted a study to analyze human performance under different conditions. For this we have asked human participants to answer a subset of $1,000$ questions from the test set given the following information:

\begin{itemize}[topsep=1pt,itemsep=1pt,partopsep=0pt, parsep=0pt, leftmargin=13pt]
    \item S1: we show the question and the image.
    \item S2: we show the question and the image but with all text instances blurred (illegible).
    \item S3: we show the question and a list of words (OCR tokens), no image is shown.
\end{itemize}

\noindent
in all three cases participants had the option to mark the questions as \textit{``unanswerable''}.

Table~\ref{tab:stvqa} shows the human performance in terms of ANLS and accuracy in the three scenarios described above. We appreciate that S1 is consistent with the human study reported in~\cite{singh2019towards} in terms of accuracy. 
Their study shows a human accuracy of $85.0$ in TextVQA, but having collected 10 answers per question their accuracy metric is a bit more flexible in accepting diverse correct answers. 
Moreover, we observe that S2 and S3 demonstrate that the textual cue is much more important than the visual cue in ST-VQA. Another point to stress is that humans are especially good at answering questions without even seeing the image. This is because of the fact that humans use a-priori knowledge of what a number is or what a licence plate is, etc. As an example, an image for which the question is \textit{``What is the price of ...''} can be correctly answered by selecting a unique numerical OCR token since the price has to be a number.


\begin{table}[h]
    \centering
    \begin{tabularx}{\linewidth}{X c c c r r}
    \toprule
      & V & T & & ANLS & Acc.\\
    \midrule
    S1 human performance & {\color{green}\cmark} & {\color{green}\cmark} & & 0.85 & 78.16 \\
    S2 human performance & {\color{green}\cmark} & {\color{red}\xmark} & & 0.21 & 18.81 \\
    S3 human performance & {\color{red}\xmark} & {\color{green}\cmark} & & 0.52 & 37.54 \\
    \bottomrule
    \end{tabularx}
    \caption{Human performance on a subset of $1,000$ questions of the ST-VQA test set under different conditions, depending whether visual~(V) or textual~(T) information is given.}
    \label{tab:human}
\end{table}

The complete results of this human study are provided as supplementary material to this paper. Furthermore, we will include in the new version of the dataset the indices of the $1,000$ test questions used in this study, and the indexes of text questions for which their answer is among the provided OCR tokens, so that interested researchers can analyze the performance of their methods on those test subsets of special interest. 

\section{Conclusion}
\label{sec:conclusion}

We have presented a new model for scene text visual question answering that is based in an attention mechanism that attends to multi-modal grid features, allowing it to reason jointly about the textual and visual modalities in the scene. 

The provided experiments and ablation study demonstrate that attending on multi-modal features is better than attending saparately to each modality. Our grid design choice also proves to work very well for this task, as well as the choice of an object detection backbone instead of a classification one. Moreover, we have shown that the proposed model is flexible enough to be combined with a standard VQA model obtaining state of the art results on mixed datasets with questions that can not be answered directly using OCR tokens.




\bibliographystyle{model2-names}
\bibliography{refs}


\end{document}